%% file: main.tex
\documentclass{vgtc}                          

\graphicspath{{figures/}{pictures/}{images/}{./}} 

\usepackage{times}                     

\usepackage{tabu}                      
\usepackage{booktabs}                  
\usepackage{lipsum}                    
\usepackage{mwe}                       
\usepackage{amsmath}
\usepackage{amssymb,algorithm, algpseudocode}
\usepackage{colortbl}

\usepackage{xspace}
\usepackage{mathtools}
\usepackage{amsthm}   

\theoremstyle{definition}

\theoremstyle{remark}

\usepackage{mathptmx}                  

\onlineid{0}

\vgtccategory{Technical papers}

\vgtcinsertpkg

\title{E-{WAVE}: \underline{E}vent-based Continuous Optical Flow \\via \underline{W}arping-\underline{A}ligned \underline{V}isual \underline{E}ncoding}

\author{%
  Jiale Wu\textsuperscript{1},
  Xiaoyang Bai\textsuperscript{2},
  Haoming Yu\textsuperscript{1},
  Yiwei Chen\textsuperscript{1},
  Yifan Peng\textsuperscript{2},
  Weiwei Xu\textsuperscript{1}%
  \thanks{%
    $^1$Zhejiang University.
    E-mail: \{jlwu\_cad, haoming\_yu, chenyiwei\}@zju.edu.cn, xww@cad.zju.edu.cn\\
    $^2$The University of Hong Kong.
    E-mail: \{xybai, evanpeng\}@hku.hk.
  }%
}

\teaser{
  \centering
  \vspace{-6pt}
  \includegraphics[width=.97\linewidth]{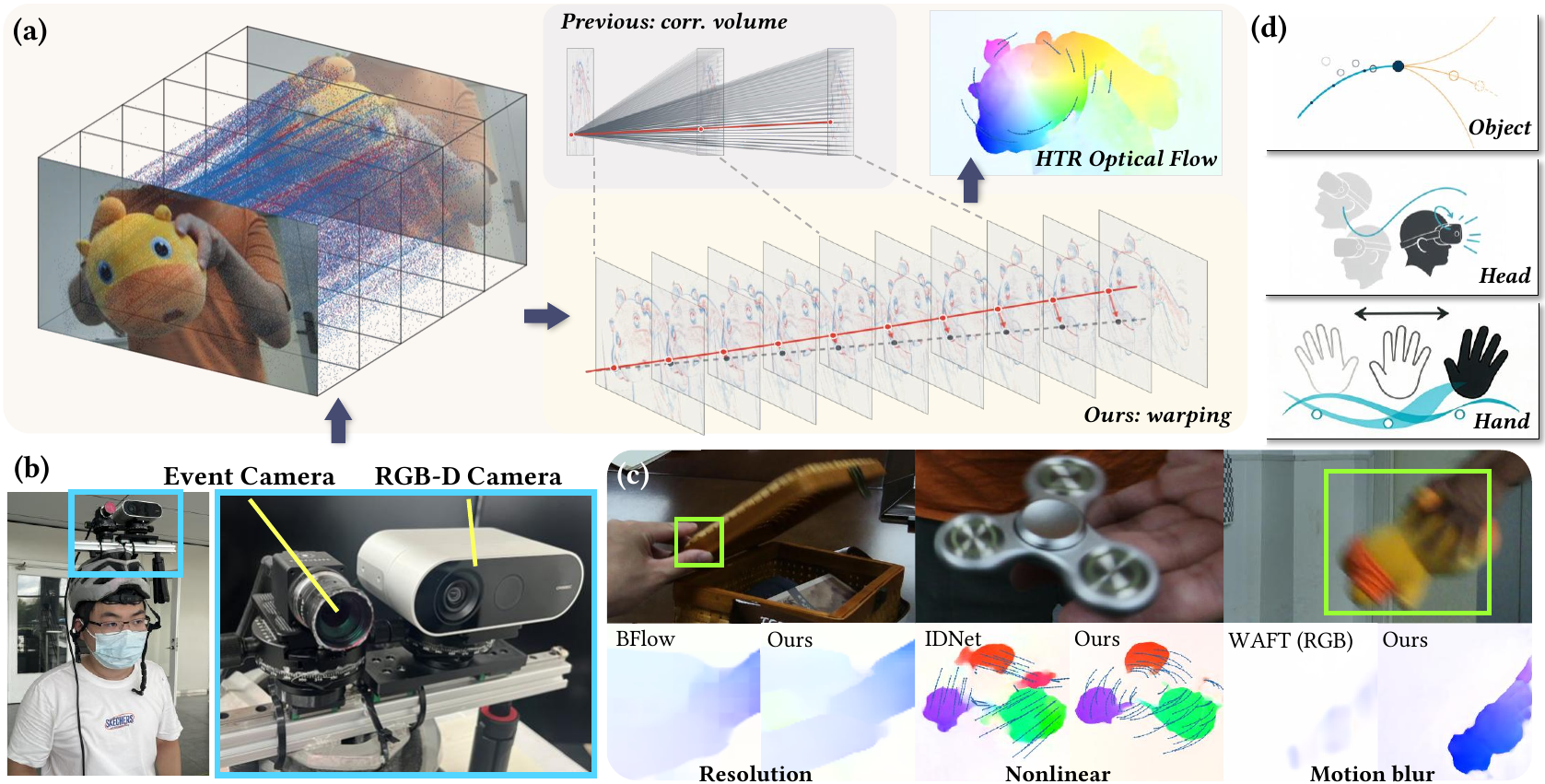}
  \vspace{-8pt}
  \caption{(a) We propose \method, an \textbf{event-based continuous optical flow estimation} method that leverages feature warping instead of correlation volumes to extract temporal and cross-modal coherence. b) Our head-mounted acquisition prototype, built to validate our model in real-world scenarios. c) \method operates at a higher resolution than existing methods such as BFlow, and it excels in challenging cases featuring nonlinear dynamics and motion blur. d) Our method is a promising candidate for modeling fast and continuous head, hand, and object motion and interaction in VR/AR applications.}
  \label{fig:teaser}
  \vspace{-2pt}
}

\input{sections/abstract}

\newcommand{\method}{E-WAVE\xspace}
\begin{document}


\input{sections/intro}

\input{sections/related}

\input{sections/method}

\input{sections/exp}

\input{sections/conclusion}



\bibliographystyle{abbrv-doi}

\bibliography{references}
\end{document}

%% file: sections/abstract.tex
\abstract{
    Temporally dense optical flow is essential for dynamic perception in immersive VR/AR systems, where rapid head, hand, and object motion must be continuously captured and tracked. 
    Existing frame-based optical flow estimation methods are constrained by the tradeoff between temporal resolution and computational cost; 
    while event cameras, with their high temporal resolution and energy efficiency, serve as a natural solution to the dilemma. However, event-based approaches commonly rely on correlation volumes to capture pairwise voxel correspondences, which incur substantial memory and computation overhead. 
    We present \method, a correlation-free framework for high-temporal-resolution (HTR) optical flow estimation from event streams. Instead of constructing all-pairs correlation volumes, \method employs global attention mechanism to model long-range feature dependencies and performs trajectory-guided feature warping using B\'ezier curve. 
    Through iterative updates, it predicts trajectories that allow for querying at arbitrary timestamps without repeated inference. 
    Experiments on MultiFlow and DSEC-Flow demonstrate a $25\%$ lower trajectory error and comparable endpoint flow estimation accuracy relative to state-of-the-art baselines. Additional evaluations on self-captured data using a head-mounted prototype validate that \method remains robust under challenging real-world conditions. 
} 

\keywords{Event Camera, Optical Flow, Motion Perception}

%% file: sections/intro.tex
\firstsection{Introduction}

\maketitle

As a crucial task in computer vision and graphics, \emph{optical flow estimation} extracts motion vectors from 2D visual data and serves as an essential prior in a wide range of downstream tasks, such as 4D reconstruction~\cite{wang2026flow4dgs}, video deblurring~\cite{Jing2022FlowvideoDeblur}, and frame interpolation~\cite{kim2023eventFrameinterp}. 
Moreover, the flourishing of real-time robotics~\cite{chao2014survey} and VR/AR~\cite{holynski2018fast} systems, which require accurate dynamic perception, also motivates the exploration of fast and robust optical flow estimation algorithms.
However, conventional frame-based approaches such as RGB cameras encounter a tradeoff between hardware capacity and estimation quality.
That is to say, temporally dense optical flow demands high frame rate inputs, increasing bandwidth requirements and computational loads, whereas lower frame rates can introduce large displacements and motion blur that degrade estimation accuracy. 
In immersive VR/AR applications, temporally dense motion estimates are essential to track fast, continuously evolving head, hand, and object interactions~\cite{banerjee2025hot3d,Han2020megatrack,Feng2020livedeepVR}, and therefore resolving this bottleneck has become a pivotal yet unresolved challenge.

In recent years, \textbf{event cameras} have emerged as a compelling tool for optical flow estimation to complement the limitations of frame-based sensors. 
By capturing per-pixel intensity changes asynchronously, event sensors offer microsecond-level temporal resolution, continuous data acquisition, high dynamic range, and immunity to motion blur~\cite{gallego2020event,lin2024embodied}. 
Exploiting these attributes, prior works prove successful in both low-temporal-resolution (LTR, \textit{i.e.}, estimating only for RGB timestamps)~\cite{gehrig2021eraft} and high-temporal-resolution (HTR, \textit{i.e.}, estimating for any arbitrary timestamp)~\cite{zhou2025resflow} regimes, even with RGB frame-free, event-only setups.
Nonetheless, state-of-the-art approaches often rely on \emph{correlation volumes} to establish feature correspondences across time and between modalities, which incurs two downsides when applied to event streams. 
Firstly, all-pairs correlation scales quadratically with spatial feature size~\cite{teed2020raft}, as it constructs a 4D volume with quadratic memory usage for every pair of feature maps. 
Even with $1/8$ spatial subsampling, the memory footprint and computation overhead can still be heavy when processing densely constructed event voxels. 
Secondly, correlation volumes only characterize pairwise correspondences and can fail to exploit long-range, continuous temporal trajectory inherently encoded by event streams.

An alternative solution to extract temporal and cross-modal coherence of features is through \textit{warping}, rather than indexing correlation volumes. 
For optical flow estimation and dense point tracking on RGB inputs, representative methods such as WAFT~\cite{wang2025waft} and CoWTracker~\cite{lai2026cowtracker} establish an effective and scalable pipeline that pairs feature warping with iterative refinement and replaces all-pairs correlation with attention-driven feature alignment. 
On the other hand, warping-based contrast maximization (CM) has also long been leveraged across diverse event-based vision tasks, such as depth estimation~\cite{gallego2018unifying} 
and feature tracking~\cite{gehrig2018asynchronous}. As an early exploration of applying warping to event-based optical flow, IDNet~\cite{Wu2024IDNet} introduces a correlation-free framework that predicts optical flow through iterative event deblurring, validating the feasibility of leveraging feature warping, albeit limited by its underlying linear motion assumption. 

Motivated by these attempts, we propose \method, an event-based, correlation-free framework that achieves \emph{HTR optical flow estimation through trajectory-guided feature warping}. 
Specifically, we leverage Vision Transformer (ViT)~\cite{Dosovitskiy2020Vit} as a replacement for correlation volumes to capture long-range dependencies, and we advance standard feature warping by parameterizing continuous pixel trajectories as B\'ezier curves. 
This design enables \method to recover temporally coherent, nonlinear motion from event streams at arbitrary timestamps. 
Analogous to state-of-the-art architectures~\cite{wang2025waft, lai2026cowtracker}, our pipeline adapts an iterative warp-and-update process for B\'ezier parameter prediction, which further extends its estimation accuracy beyond single-step inference.

Experimental results show that \method achieves state-of-the-art performance when estimating HTR optical flow on the challenging MultiFlow dataset~\cite{gehrig2024bflow} and can be extended with multimodal RGB inputs, bidirectional feature matching, and cross-dataset finetuning, yielding substantial improvements on the LTR DSEC-Flow benchmark~\cite{Gehrig21DSEC}. 
To evaluate real-world robustness, we build a headset prototype with stereo event-RGB cameras and test on self-captured data featuring challenging cases such as underexposure and non-linear motion. 
These field evaluations reveal that \method reliably handles complex motion patterns induced by egocentric capturing with typical head-mounted or hand-held devices, highlighting its strong potential for immersive VR/AR applications.
In all, our main contributions are summarized as follows:
\vspace{-1pt}
\begin{itemize}
    \item We propose \emph{\method}, a trajectory-guided warping framework for event-based HTR optical flow estimation, that achieves state-of-the-art performance on the challenging MultiFlow dataset while maintaining competitive LTR capability. Moreover, extended versions of our model achieve state-of-the-art on LTR DSEC-Flow benchmark.
    \item We design a unified, flexible feature construction, warping and updating pipeline for both event-only and event-RGB setups. Given this, one can easily customize the pipeline across diverse input modalities and motion dynamics. 
    \item Besides extensive benchmark evaluations, we also validate on real-world sequences captured by a head-mounted acquisition prototype, demonstrating its robustness and applicability to VR/AR scenarios.
\end{itemize}
\vspace{-2pt}

%% file: sections/related.tex
\section{Related Work}
\subsection{Optical Flow Estimation}

Deep learning methods for optical flow estimation has witnessed great progress in recent years since the pioneering FlowNet~\cite{Dosovitskiy2015flownet}. Follow-up works such as PWC-Net~\cite{Sun2018PWC-Net} improve estimation quality with components including pyramid processing, warping, and correlation-volume construction. 
The RAFT series~\cite{teed2020raft,wang2024searaft} introduce iterative refinement and demonstrate superior performance and efficiency, as well as the effectiveness of correlation volumes on handling large displacements.
To avoid quadratic computational complexity of 4D correlation volumes, subsequent works adopt partial correlation volume~\cite{morimitsu2025dpflow} or global attention mechanism~\cite{xu2022gmflow}. 
Most recently, WAFT~\cite{wang2025waft} further combines warping with attention mechanism to estimate optical flow at a higher resolution, as will be discussed in detail below.

\subsection{Event-based Optical Flow Estimation}

Early approaches exploit event signals with hand-crafted heuristics~\cite{Benosman2012async,brosch2015evbasedof}, typically under restrictive assumptions that limit their applicability to real-world scenarios~\cite{dalgaty2023hugnet,Zhu2018evflownet}. 
Later, the principle of Contrast Maximization is introduced to tackle various applications~\cite{gallego2018unifying}, optical flow estimation included. This formulation better reflects intrinsic characteristics of events by aligning moving edges to one or multiple timestamps via warping~\cite{shiba2024secrets}, 
Alternative works design specialized loss functions and supervise learning-based methods~\cite{hagenaars2021selfsup,hamann2024motion,hu2026stsc}, but they can struggle with the inherently sparse and noisy nature of event signals~\cite{shiba2022eventcollapse,han2025edef}. 

The availability of large-scale datasets~\cite{li2023blinkflow}, built through simulation~\cite{Bai2026eventtracer,hu2021v2e} or dedicated acquisition~\cite{delaney2025evaluating}, has facilitated recent data-driven methods. 
The pioneering E-RAFT~\cite{gehrig2021eraft}, inspired by RAFT~\cite{teed2020raft}, introduces \textit{iterative refinement with correlation volumes}, supported by discrete event voxels~\cite{Zhu2019EvVoxel}. 
To further capture intermediate motion, TMA~\cite{liu2023tma} constructs multiple correlation volumes within each interval, a strategy later adopted in RGB- or stereo-augmented approaches~\cite{zhou2026STFlow,zhang2025ematch}. BAT~\cite{Xu2026BAT} extends this idea in a bidirectional manner, achieving remarkable results. 
However, the quadratically scaling memory usage of correlation volumes severely limits spatial resolution. To this end, IDNet~\cite{Wu2024IDNet} alternates the process with iterative deblurring, while EDC-Flow~\cite{liu2025edcflow} combines feature differences on higher resolution with correlation volumes for improved estimation.

In HTR optical flow estimation, DCEIFlow~\cite{wan2022dcei} constructs pseudo-next-frame features for dense correlation volumes to predict intermediate optical flow. 
BFlow~\cite{gehrig2024bflow} replaces linear motion assumptions with correlation sampling guided by predicted B\'ezier curves~\cite{seok2020quadbezier}. 
ResFlow~\cite{zhou2025resflow} and STSC-Flow~\cite{hu2026stsc} incorporate extra supervision to learn HTR trajectories with LTR ground truth. 
However, scalability remains a fundamental limitation: these approaches must subsample temporally due to the burden brought by correlation volumes.
Similar challenges arise in event-based point tracking tasks~\cite{wan2025edcpt}. 
Alternatively, EVA-Flow~\cite{ye2025evaflow} explores a correlation-free framework to predict instant velocity, and then predict final optical flow via accumulated velocities, though drifting error progressively degrades performance on longer sequences.

In summary, although correlation volumes empower event-based optical flow estimation with higher accuracy, their memory-expensive nature systematically limits the scalability of relevant techniques towards higher resolution, both spatially and temporally. 

\subsection{Feature Alignment with Warping}

Warping has been widely used by classical and early learning-based optical flow estimation methods across RGB and event inputs~\cite{Ranjan2017pyramid, paredes2021backbasics}, where it is considered as a tool for photometric brightness alignment or contrast maximization~\cite{paredes2023TCM}. 
Following PWC-Net~\cite{Sun2018PWC-Net}, SCFlow~\cite{hu2022scflow} warps feature in a pyramidal style and construct correlation volumes after warping. This reveals that aligned features can serve as tokens for feature matching. Recently, WAFT~\cite{wang2025waft} further combines iterative warping with global attention mechanism for effective optical flow estimation, and CoWTracker~\cite{lai2026cowtracker} similarly utilizes attention mechanism to track spatially large motion. Additionally, SWIFT~\cite{wang2026swift} combines warping with correlation volumes in lower resolution for better efficiency.

Existing warping-based approaches on event cameras~\cite{Wu2024IDNet,ye2025evaflow} often assume short temporal interval and constant motion for input data, but these assumptions may not hold for real-world applications on immersive systems, where accelerated, articulated, or otherwise nonlinear motion~\cite{Friedhelm2025ETAP,li2023blinkflow} commonly occurs. In such cases, endpoint flow prediction with linear warping may fail to align motion between frames~\cite{seok2020quadbezier,gehrig2024bflow}, whereas how to unleash warping without the linear-motion assumption for optical flow estimation is still under-explored. To address this limitation, we extend warping to continuous event-guided trajectory represented as B\'ezier displacement fields, which allows warping to account for nonlinear motion in event-based scenarios.

\begin{figure}
\centering
  \includegraphics[width=0.99\linewidth]{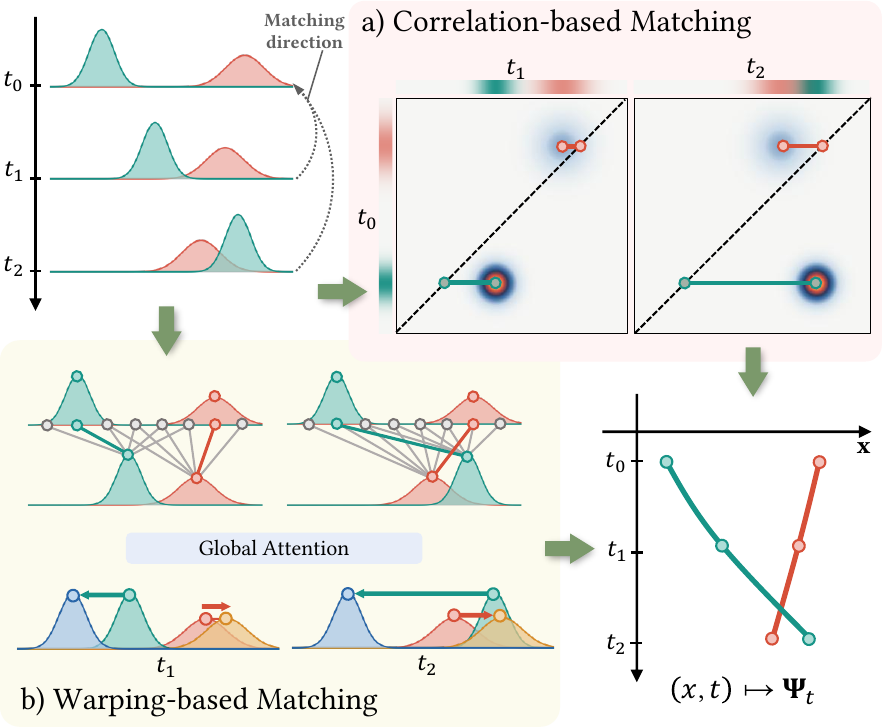}
  \caption{Correlation-based \textit{vs.} Warping-based matching. (a) Correlation volumes store dense pixel-wise similarity between two feature frames, which is queried at a later stage to update estimated trajectories. (b) Our method leverages global attention to capture long-range correlation and directly derive warping parameters.}
  \vspace{-3pt}
  \label{fig:bezeir warping}
\end{figure}

\begin{figure*}[ht]
\centering
  \includegraphics[width=0.995\linewidth]{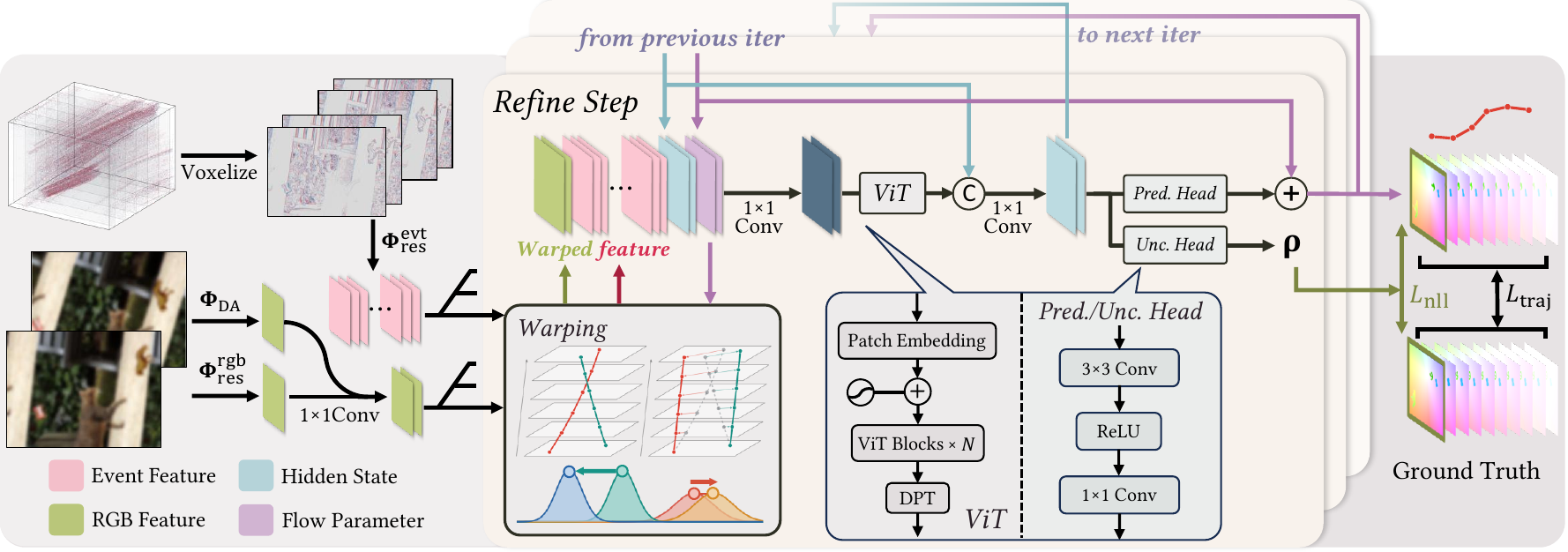}
  \caption{Pipeline of \method. 
  We firstly extract features from voxelized events and RGB frames (if available), and then repeat the refinement step by $R$ times. At each iteration, visual features are warped using the predicted trajectory map from the previous iteration (purple), concatenated with the trajectory map and the hidden state (cyan), and processed by a ViT-DPT module to obtain an updated hidden state. Ultimately, two convolution heads convert the hidden state into the updated trajectory map and uncertainty vector $\rho$, respectively. An NLL loss on endpoints and a trajectory loss on intermediate ground truth are computed for each iteration to train the entire framework.}
  \label{fig:pipeline}
\end{figure*}

%% file: sections/method.tex
\section{\method Method}

\label{sec:method}


\subsection{Event Camera and Event Voxelization}
Unlike conventional sensory, event cameras asynchronously trigger events in response to changes in logarithmic brightness $L=\log I$, producing a sparse and variable-length event stream instead of regularly sampled image tensors~\cite{Zhu2018evflownet}. An event activation is fired when $\Delta L\approx pC$, where $C$ is the contrast threshold and $p$ is the event polarity (+1 or -1). Under the brightness-constancy assumption, local event rate $r$ is determined jointly by the spatial intensity gradient $\nabla L$ and image motion $\mathbf{v}$:
\begin{equation}
    6,
\end{equation}
and therefore event signal serves as a good motion indicator.

However, directly tensorizing raw events leads to inconsistent input dimensions and large memory overhead. The conventional practice is thus to convert event streams into compact, fixed-dimensional voxel representations~\cite{zhou2026STFlow}. Specifically, given bin width $\Delta t$ and a set of anchor timestamps $\{\tau_n\}$, $n\in 0,1,\ldots,K$ between two adjacent RGB frames, we define the spatiotemporal interpolation weight for an event $e_i=(x_i,y_i,t_i,p_i)$ at pixel $(x, y)$ as:
\begin{equation}
\begin{aligned}
    w_n(x,y)
    &=b(x_i-x)b(y_i-y)
      b\!\left(\frac{t_i-\tau_n}{\Delta t}\right),\\
   \text{and} \;\;  b(u)&=\max(0,1-|u|).
\end{aligned}
\label{eq:fine_event_weight}
\end{equation}
Then we accumulate the contribution of all events at each pixel, for positive and negative activations separately:
\begin{equation}
    V_n^s(x,y)
    =\sum_{i:\,p_i=s} w_n(x,y),
    \qquad s\in\{+1,-1\}.
    \label{eq:polarity_fine_voxels}
\end{equation}
To align with the dimension of RGB images, we construct three-channel event voxels by additionally computing the difference between two polarity channels after clipping and normalization:
\begin{equation}
    \mathbf{V}_n
    =\bigl[V_n^{+},\,V_n^{-},\,V_n^{+}-V_n^{-}\bigr]
    \in\mathbb{R}^{3\times H\times W}.
    \label{eq:three_channel_fine_voxel}
\end{equation}

\subsection{Problem Definition and Feature Matching}

Given the constructed event voxels $\{\mathbf{V}_n\}_{n=0}^K$ between two boundary timestamps (endpoints), $\tau_0$ and $\tau_K$, and optionally the corresponding RGB frames $\mathbf{I}_0$ and $\mathbf{I}_1$, \textit{low-temporal-resolution (LTR)} optical flow estimation aims to retrieve a pixelwise flow map $\boldsymbol{\Psi}$ from $\tau_0$ to $\tau_K$ (\textit{i.e.}, from $\mathbf{I}_0$ to $\mathbf{I}_1$), while \textit{high-temporal-resolution (HTR)} estimation is able to compute flow map $\boldsymbol{\Psi}_t$ corresponding to any intermediate timestamp $t \in [\tau_0,\tau_{K}]$.


As scene motion inevitably causes spatial misalignment between voxels (and frames), a crucial step in optical flow estimation is \textit{feature matching}. Given feature maps $\mathbf{F}^{a}$ and $\mathbf{F}^{b}$, the classic \textbf{correlation-based approach} constructs a correlation volume $\mathbf{C}^{a,b}$ that stores similarities over candidate displacements $\mathcal{D}$:
\begin{equation}
    \mathbf{C}^{a,b}(\mathbf{x},\mathbf{d})
    =
    \left\langle
        \mathbf{F}^{a}(\mathbf{x}),
        \mathbf{F}^{b}(\mathbf{x}+\mathbf{d})
    \right\rangle,\quad \mathbf{d}\in\mathcal{D}.
\end{equation}
In subsequent steps, $\mathbf{C}^{a,b}$ is queried by estimated flow maps to retrieve local correspondence information along each pixel trajectory. Alternatively, \textbf{warping} computes a per-pixel displacement matrix $\mathbf{U}^{a,b}$ and aligns two features using grid sampling:
\begin{equation}
    \widetilde{\mathbf{F}}^{b\rightarrow a}(\mathbf{x})
    \coloneqq
    \mathbf{F}^{b}\left(
    \mathbf{x}+\mathbf{U}^{a,b}(\mathbf{x})
    \right).
    \label{eq:warp}
\end{equation}
Correlation relies on a separate matching hypothesis for each feature pair, whereas warping produces a single aligned feature map for a whole feature sequence. The latter design therefore allows for higher spatial and temporal resolution and subsequently leads to better performance, especially when operating on densely voxelized event data. See Fig.~\ref{fig:bezeir warping} for a visual comparison.

\subsection{Warping-Aligned Visual Encoding}
\label{sec:warping_htr}
As illustrated in Fig.~\ref{fig:pipeline}, our \method pipeline consists of three steps: (a) extracting multimodal features via RGB and event voxel encoders, (b) warping features with a parametrized displacement field, and (c) predicting residual updates using warped features with a global attention mechanism. Steps (b) and (c) are recursively applied for $R$ iterations to yield the final output.

\subsubsection{Feature extraction and fusion}

To maintain architectural consistency across modalities, we employ the same ResNet~\cite{He2016Resnet} architecture for both RGB frame encoder $\Phi_{\mathrm{res}}^{\mathrm{rgb}}$ and event voxel encoder $\Phi_{\mathrm{res}}^{\mathrm{evt}}$, which differ by only the last projection layer. We further complement RGB frame features by depth-aware features extracted from a frozen Depth Anything V2~\cite{yang2024dav2} network, and then project them to the same dimension as event features via a standalone $1\times1$ convolution. The final per-frame RGB features $\mathbf{R}_0, \mathbf{R}_1$ and per-voxel event features $\mathbf{F}_j$, are formally written as:
\begin{equation}
\begin{aligned}
    \mathbf{R}_i
    &=\operatorname{Conv}_{1\times1}\left(\Big[
      \Phi_{\mathrm{DAv2}}(\mathbf{I}_i),\Phi_{\mathrm{res}}^{\mathrm{rgb}}(\mathbf{I}_i)\Big]\right),
      && i\in\{0,1\},\\
    \mathbf{F}_j
    &=\Phi_{\mathrm{res}}^{\mathrm{evt}}(\mathbf{V}_j),
      && j=0,\ldots,K.
\end{aligned}
\end{equation}
We also apply a shared 2D positional encoding to each event feature map before feature warping.

\subsubsection{Trajectory parameterization}
\label{sec:traj_parameterization}
Throughout the estimation process, we maintain a parametrized displacement field $\mathbf{U}$, where $\mathbf{U}(\mathbf{x}, t)$ denote the displacement of pixel $\mathbf{x}$ at time $t\in[\tau_0, \tau_K]$ relative to $\tau_0$. 
At each pixel location $\mathbf{x}$ (henceforth omitted for clarity), $\mathbf{u}(t)\coloneqq \mathbf{U}(\mathbf{x}, t)$ is represented as a degree-$L$ B\'ezier curve with anchors $\mathbf{p}\in\mathbb{R}^L$. Formally:
\begin{equation}
    \mathbf{u}(t)
    =
    \sum_{\ell=0}^{L}
    \beta_{\ell}^{L}(\hat{t})\mathbf{p}_{\ell},
    \qquad
    \beta_{\ell}^{L}(\hat{t})
    =
    {L\choose\ell}(1-\hat{t})^{L-\ell}\hat{t}^{\ell}.
    \label{eq:bezier_trajectory}
\end{equation}
Here $\hat{t}$ is normalized time defined as $\hat{t} = (t - \tau_0) / (\tau_K - \tau_0)$.

Since the trajectory represents displacement
relative to $\tau_0$, we set
$\mathbf{p}_0=0$, which ensures that
$\mathbf{u}(\tau_0)=0$.
Instead of directly predicting $\mathbf{p}$, our network equivalently outputs $\boldsymbol{\theta}$ as differences between adjacent
control points, that is,
\begin{equation}
    \mathbf{p}_{\ell}
    =
    \sum_{q=1}^{\ell}\boldsymbol{\theta}_q,
    \qquad
    \ell=1,\ldots,L.
    \label{eq:control_from_increments}
\end{equation}
We use $\boldsymbol{\Theta}$ to denote the collection of $\boldsymbol{\theta}$'s over all pixel locations.

\subsubsection{Feature warping and residual update}
At the $r$-th iteration, we warp $\{\mathbf{F}_j\}$, $\mathbf{R}_0$ and $\mathbf{R}_1$ (if available) using the up-to-date $\mathbf{U}^{(r)}$ to obtain aligned features:
\begin{equation}
    \widetilde{\mathbf{F}}_j^{(r)}\gets \operatorname{warp}(\mathbf{F}_j,\mathbf{U}^{(r)}(\cdot, \tau_j)), \;\text{and} \;\;
    \widetilde{\mathbf{R}}_1^{(r)}\gets \operatorname{warp}(\mathbf{R}_1,\mathbf{U}^{(r)}(\cdot, \tau_K)).
\label{eq:warpwarp}
\end{equation}
Here $\operatorname{warp}(\cdot)$ is defined as applying Eq.~\ref{eq:warp} for each pixel. We initialize $\mathbf{U}^{0}$ to be all-zeros, and subsequent $\mathbf{U}^{(r)}$'s are parametrized by estimated control parameters $\boldsymbol{\Theta}^{(r)}\in\mathbb{R}^{L\times H\times W}$, as detailed above.

Then, we assemble the latent feature map $\mathbf{z}$ from aligned event features $\widetilde{\mathbf{F}}_j$, reference image features $\mathbf{R}_0$ and $\widetilde{\mathbf{R}}_1$, hidden recurrent state $\mathbf{h}$, and current control parameters $\boldsymbol{\Theta}$ using $1\times1$ convolution:
\begin{equation}
    \mathbf{z}^{(r)} =\operatorname{Conv}_{1\times1}\left(\Big[
        \widetilde{\mathbf{F}}_0^{(r)},\ldots,
        \widetilde{\mathbf{F}}_K^{(r)},
        \mathbf{R}_0,\widetilde{\mathbf{R}}_1^{(r)},
        \mathbf{h}^{(r)},\operatorname{sg}[\boldsymbol{\Theta}^{(r)}]\Big]\right).
\label{eq:refinement_input}
\end{equation}
$\mathbf{R}_0$ and $\widetilde{\mathbf{R}}_1$ are set to zero if RGB inputs are unavailable. Next, we pass $\mathbf{z}$ through a ViT-DPT~\cite{Ranftil2021DPT} refinement network $T_\mathrm{ref}$, which embeds $\mathbf{z}$ using an \(8\times8\) patch projection, adds an interpolated learnable positional embedding, and decodes intermediate transformer features into a dense output on the spatial grid, that is, $\mathbf{y}^{(r)}=T_{\mathrm{ref}}(\mathbf{z}^{(r)})$. We then update the hidden state by:
\begin{equation}
\begin{aligned}
    \mathbf{h}^{(r+1)}
    &=\operatorname{Conv}_{1\times1}\!\left(
      \Big[
        \mathbf{y}^{(r)},\mathbf{h}^{(r)}\Big]\right).
\end{aligned}
\label{eq:hidden_refinement}
\end{equation}

\begin{algorithm}[t]
    \caption{Iterative trajectory refinement.}
    \label{alg:trajectory_warping}
    \begin{algorithmic}[1]
        \Require Event voxel features $\{\mathbf{F}_j\}_{j=0}^{K}$; RGB image features
        $\mathbf{R}_0,\mathbf{R}_1$; initial state
        $\mathbf{h}^{(0)}$; number of iterations $R$.
        \State $\mathbf{U}^{(0)}\gets\mathbf{0}$
        \For{$r=0,\ldots,R-1$}
            \State $\widetilde{\mathbf{F}}_j^{(r)} \gets \mathbf{F}_j, \mathbf{U}^{(r)}$;\hspace{3pt} $\widetilde{\mathbf{R}}_1^{(r)} \gets \mathbf{R}_1, \mathbf{U}^{(r)}$ \hfill(Eq.~\ref{eq:warpwarp}) 
            \State $\mathbf{z}^{(r)}\gets \widetilde{\mathbf{F}}_0^{(r)},\ldots,
            \widetilde{\mathbf{F}}_K^{(r)},
            \mathbf{R}_0,\widetilde{\mathbf{R}}_1^{(r)},
            \mathbf{h}^{(r)},
            \mathrm{sg}[\boldsymbol{\Theta}^{(r)}]$ \hfill(Eq.~\ref{eq:refinement_input})
            \State $\mathbf{h}^{(r+1)}\gets T_{\mathrm{ref}}(\mathbf{z}^{(r)}), \mathbf{h}^{(r)}$ \hfill(Eq.~\ref{eq:hidden_refinement})
            \State $\delta\boldsymbol{\Theta}^{(r)}\gets \mathbf{h}^{(r+1)}$ \hfill(Eq.~\ref{eq:control_head})
            \State $\mathbf{U}^{(r+1)} \gets B(\delta\boldsymbol{\Theta}^{(r)})$ \hfill(Eq.~\ref{eq:control_refinement} \& Sec.~\ref{sec:traj_parameterization})
        \EndFor
        \State \Return $\boldsymbol{\Psi}_t\coloneqq\mathbf{U}^{(R)}$
    \end{algorithmic}
    \vspace{-3pt}
\end{algorithm}

\begin{figure*}
\centering
  \includegraphics[width=0.99\linewidth]{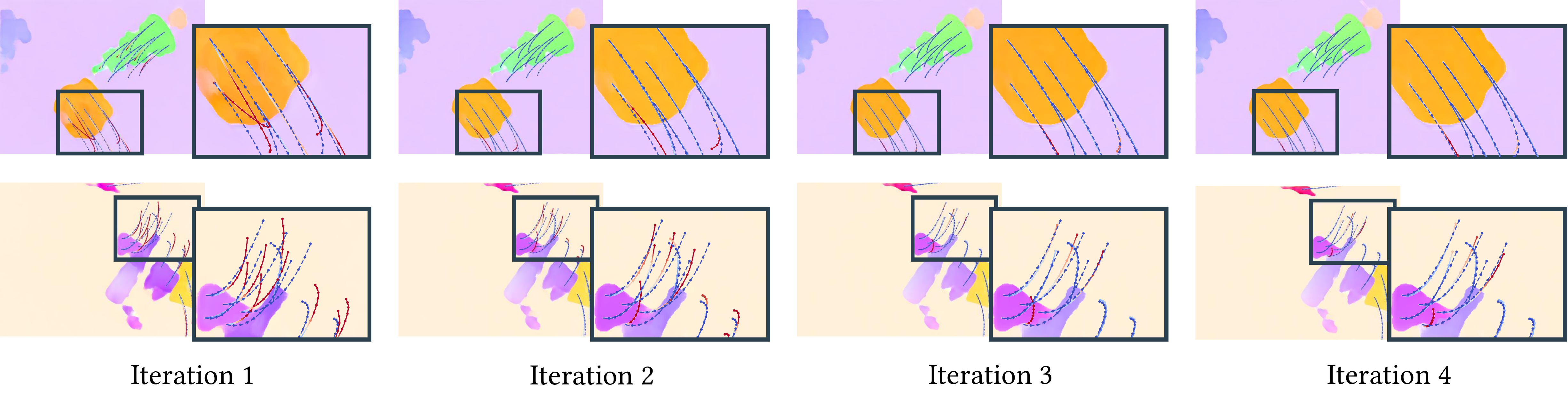}
  \vspace{-9pt}
  \caption{Visualization of estimated optical flow and predicted (red) \textit{vs.} ground truth (blue) trajectories across different iterations. The warp-and-update process improves both pixelwise trajectory alignment and object-level flow coherence in the first few iterations and converges afterwards.}
  \label{fig:bezeir_linear}
  \vspace{-3pt}
\end{figure*}

Finally, we use a prediction head to output increments of control parameters and an uncertainty head for NLL loss computation, each with two convolution layers:
\begin{equation}
\begin{aligned}
\delta\boldsymbol{\Theta}^{(r)}&=\operatorname{Conv}_{1\times1}\left[\operatorname{ReLU}\left(\operatorname{Conv}_{3\times3}
      (\mathbf{h}^{(r+1)})\right)\right],\\
\boldsymbol{\rho}^{(r)}&=\operatorname{Conv}_{1\times1}\left[
      \operatorname{ReLU}\left(
      \operatorname{Conv}_{3\times3}
      (\mathbf{h}^{(r+1)})\right)\right].\\
\label{eq:control_head}
\end{aligned}
\end{equation}
Refer to Sec.~\ref{sec:training_objective} for more details on NLL loss. With the estimated $\delta\boldsymbol{\Theta}^{(r)}$, we may then update the control parameters as:
\begin{equation}
\begin{aligned}
    \Theta^{(r+1)}
        &=\operatorname{sg}[\Theta^{(r)}]
          +\delta\Theta^{(r)},
\end{aligned}
\label{eq:control_refinement}
\end{equation}
and proceed into the next iteration. Algorithm~\ref{alg:trajectory_warping} summarizes the complete iterative refinement procedure. We use $B(\delta\boldsymbol{\Theta})$ to denote the composition of Eq.~\ref{eq:control_refinement} and B\'ezier curve construction in Sec.~\ref{sec:traj_parameterization}. Note that we operate on a $2\times 2$ downsampled scale to reduce memory and time consumption, and a learned convex $3\times 3$ kernel is used to upsample $\boldsymbol{\Theta}^R$ back to the input resolution before outputting it as $\boldsymbol{\Psi}_t$.

\subsection{Analysis on Iterative Warp-and-Update}

\method's \textit{alternating warp-and-update process} allows it to model long-range correlations and dependencies that are hard to capture with one-step inference, even with global attention. To further analyze its optimization dynamics, observe that Eq.~\ref{eq:warp}, Eq.~\ref{eq:bezier_trajectory}, Eq.~\ref{eq:control_from_increments} and Eq.~\ref{eq:control_refinement} are all linear. Therefore, after $r$ iterations, the trajectory maps and warped features can be written as:
\begin{equation}
    \begin{aligned}
        \mathbf{U}^{(r+1)} &= \sum_{i=0}^rB(\delta\boldsymbol{\Theta}^{(i)}) = B(\sum_{i=0}^r\delta\boldsymbol{\Theta}^{(i)}),\\
        \widetilde{\mathbf{F}}^{(r+1)} &= \operatorname{warp}(\mathbf{F}, \mathbf{U}^{(r+1)}) = \operatorname{warp}(\mathbf{F}, B(\sum_{i=0}^r\delta\boldsymbol{\Theta}^{(i)})).
    \end{aligned}
\end{equation}
That is, if we define $\bar{\boldsymbol{\Theta}}\coloneqq\sum_{i=0}^r\delta\boldsymbol{\Theta}^{(i)}$, then the first $r$ iterations are equivalent to one single step of inference with the refinement network $T_\mathrm{ref}$ predicting $\bar{\boldsymbol{\Theta}}$. In other words, increasing the number of iterations does not inherently cause the output to diverge.

However, this property does not guarantee iterative convergence, which by itself is hard to prove theoretically due to the nonconvex nature of $T_\mathrm{ref}$. To empirically assess whether \method behaves as expected across iterations, we visualize its per-iteration outputs in Fig.~\ref{fig:bezeir_linear}. As shown in the figure, estimation quality improves significantly for the initial iterations and stabilizes afterwards, which supports our intuitive assumption that warp-and-update converges to a well-matched state. 

\subsection{Training Objectives}
\label{sec:training_objective}


We supervise every refinement iteration with an endpoint uncertainty loss and, when temporally dense annotations are available, a trajectory loss. For the endpoint loss, we follow WAFT~\cite{wang2025waft} and SEA-RAFT~\cite{wang2024searaft} to use a mixture-of-Laplace formulation. At iteration $r$, the prediction head (Eq.~\ref{eq:control_head}) maps the refined hidden state to $\rho\in\mathbb{R}^{H\times W\times 4}$, which represents two mixture logits and two log-scales per pixel. Denote the ground truth optical flow as $\mathbf{U}^*$, the endpoint residual $e$ at pixel $\mathbf{x}$ can be written as:
\begin{equation}
    e = \mathbf{U}^*(\mathbf{x}, \tau_K) - \mathbf{U}^{(r)}(\mathbf{x}, \tau_K),
\end{equation}
and we model $e$ using a two-component Laplace mixture and minimize the negative log-likelihood of the ground-truth flow:
\begin{align}
    &p(e)=\alpha\frac{\exp(-|e|)}{2}
+(1-\alpha)\frac{\exp\!\left(-|e|/\exp(\beta)\right)}{2\exp(\beta)},\\
&\ell_{\mathrm{NLL}}(e)=-\log p(e),
\end{align}
where $\alpha$ and $\beta$ are calculated from $\rho(\mathbf{x})$. The final loss $L_{\mathrm{nll}}^{(r)}$  averages \(\ell_{\mathrm{NLL}}\) over all valid pixels and two flow directions. This loss respects the behavior of the standard \(L_1\) objective while taking into account large uncertainty in ambiguous or heavily occluded regions. That is, unpredictable samples are less likely to dominate optimization, reducing overfitting in presence of ambiguous cases and improving cross-dataset generalizability.

For dense trajectory supervision, let $\Omega_q$ be the set of valid pixels at normalized time $\tau_q$, we evaluate the predicted B\'ezier curve using a vector Charbonnier penalty:
\begin{equation}
\begin{aligned}
    L_{\mathrm{traj}}^{(r)}
    =\frac{1}{|\mathcal{T}|}
      \sum_{q\in\mathcal{T}}
      \frac{1}{|\Omega_q|}
      \sum_{\mathbf{x}\in\Omega_q}
      \left(
      \sqrt{\left\|
        \mathbf{U}^{(r)}(\mathbf{x}, \tau_q)
        -\mathbf{U}^{*}(\mathbf{x}, \tau_q)
      \right\|_2^2+\varepsilon^2}
      -\varepsilon
      \right),
\end{aligned}
\label{eq:trajectory_loss}
\end{equation}
where $\mathcal{T}=\{q:|\Omega_q|>0\}$. Note that averaging each timestamp by $\Omega_q$ prevents timestamps with more valid pixels from dominating loss calculation.

Finally, we exponentially weight both loss terms from all $R$ refinement iterations to obtain the overall loss value:
\begin{equation}
    L
    =\sum_{r=1}^{R}\gamma^{R-r}
      \left(
        L_{\mathrm{nll}}^{(r)}
        +\lambda_{\mathrm{traj}}L_{\mathrm{traj}}^{(r)}
      \right).
\label{eq:sequence_loss}
\end{equation}

\subsection{Extensions for Improved Performance}
\label{sec:extension}
While \method adopts a simple baseline configuration by default, its warping-based design can be readily extended with techniques commonly used in existing optical-flow methods. Specifically, we extend its capacity by incorporating \textit{bidirectional feature matching} and \textit{cross-dataset finetuning}, which improve model performance without altering the core warping-based framework.

\begin{figure*}[ht]
\centering
  \includegraphics[width=0.98\linewidth]{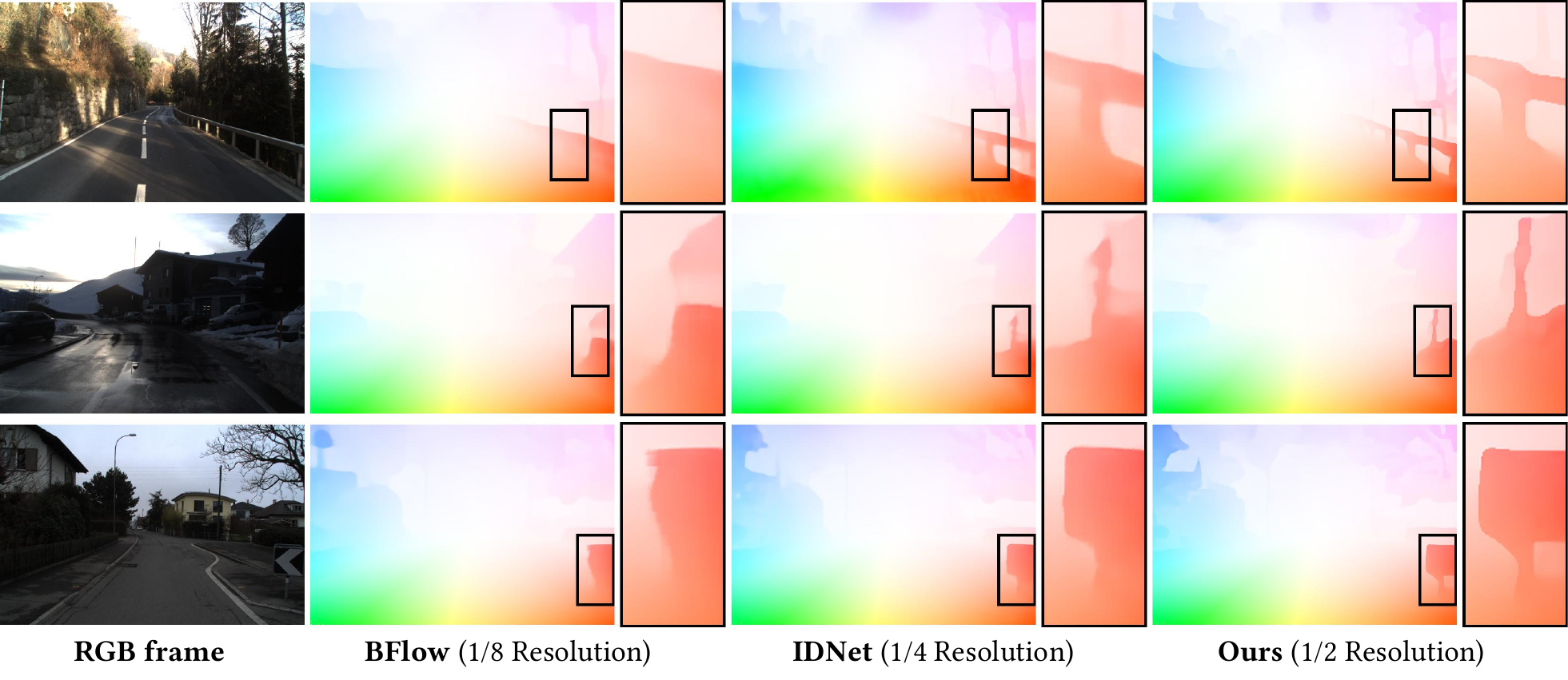}
  \vspace{-9pt}
  \caption{Qualitative comparison on DSEC-Flow~\cite{gehrig2021eraft}. \method produces more detailed estimation thanks to its higher spatial resolution. We cropped bottom 60px of the figure for the lack of supervision in this region.}
  \label{fig:dsec_compare}
\end{figure*}

\textbf{Bidirectional feature matching} has been explored in both RGB-based~\cite{Shi2023videoflow} and event-based~\cite{Xu2026BAT} approaches. Specifically, we extend \method to jointly predict the forward and backward endpoint flows $\mathbf{U}_{\mathrm{fwd}}$ and $\mathbf{U}_{\mathrm{bwd}}$. They are then used to construct a degree-2 motion trajectory:
\begin{equation}
    \mathbf{U}(\tau) =
    \frac{1}{2}
    \left(\mathbf{U}_{\mathrm{bwd}}+\mathbf{U}_{\mathrm{fwd}}\right)\tau^2
    +
    \frac{1}{2}
    \left(\mathbf{U}_{\mathrm{fwd}}-\mathbf{U}_{\mathrm{bwd}}\right)\tau,
\end{equation}
where $\tau\in[-1,1]$ denotes the normalized timestamp. Event voxel features are warped along the resulting trajectory in both temporal directions, while the core encoding and refinement modules remain unchanged. The forward and backward flows are jointly optimized using the corresponding bidirectional ground truth available in datasets such as DSEC-Flow~\cite{gehrig2021eraft}. 

\textbf{Cross-dataset finetuning} follows the strategy adopted by ECDPT~\cite{wan2025edcpt} and STFlow~\cite{zhou2026STFlow}, where we firstly pretrain \method on a larger synthetic dataset (\textit{e.g.} MultiFlow~\cite{gehrig2024bflow}) and then finetune it on real-world data (\textit{e.g.} DSEC-Flow). This strategy complements the spatially sparse annotation of real-world datasets and mitigates overfitting observed in our experiments.

%% file: sections/exp.tex
\section{Experiments and Results}



\subsection{Datasets}
We conduct extensive experiments on DSEC-Flow~\cite{gehrig2021eraft,Gehrig21DSEC} and MultiFlow~\cite{gehrig2024bflow}. DSEC-Flow is a well-established event-RGB driving dataset with $10$ Hz ground truth annotations. Although it lacks HTR optical flows and therefore can only be used for evaluating LTR performance, we still conduct relevant training and validation to quantitatively assess our model's performance in real-world scenarios. Predictions on the held-out test set are submitted to the official online evaluation server for evaluation. 

MultiFlow is a more challenging synthetic dataset featuring dynamic, nonlinear motion patterns. It serves as the primary benchmark for evaluation continuous motion trajectories estimated by HTR methods. Each sample from the dataset contains dense ground-truth trajectories from $\tau_0=400$ ms to $\tau_K=900$ ms, with reference RGB frame rendered at $\tau_0=400$ ms. Due to storage constraints, we train our model on a subset of $2{,}000$ samples, approximately one-fifth of the full training set, while evaluating on the complete test set. 

\subsection{Baselines}
For DSEC-Flow, we compare against representative baseline methods on the official leaderboard, including E-RAFT~\cite{gehrig2021eraft}, TMA~\cite{liu2023tma}, IDNet~\cite{Wu2024IDNet}, ECDDP~\cite{yang2024event}, BAT~\cite{Xu2026BAT}, ResFlow~\cite{zhou2025resflow}, BFlow~\cite{gehrig2024bflow}, and STFlow~\cite{zhou2026STFlow}. We directly use their reported metric scores on the official benchmark website for comparison. 
For MultiFlow, we retrain a wide range of open-sourced methods by ourselves, covering different input modalities:

\input{exps/dsec}

\vspace{2pt}
\noindent\textbf{Event-only methods}. IDNet, E-RAFT, TMA, BFlow, and ResFlow are chosen as baselines. IDNet, E-RAFT, and TMA are LTR methods that produce only endpoint flow estimates, and we compute their intermediate trajectories by linearly interpolating two endpoint predictions. For the two baseline HTR methods, BFlow and ResFlow, as well as our \method, we use the same degree of freedom for trajectory representation. Specifically, we set the degree of B\'ezier curves to 10 in \method and BFlow, and train ResFlow to predict 10 dense flow fields at uniformly sampled timestamps too.

\vspace{2pt}
\noindent\textbf{Event-RGB method}. We compare against the state-of-the-art STFlow in this category. Likewise, we extend STFlow to predict degree-10 B\'ezier curves to align with other HTR methods.

\vspace{2pt}
\noindent\textbf{RGB-only method}. We additionally train WAFT~\cite{wang2025waft} on RGB frames only for a more thorough comparison.

\subsection{Metrics}

We evaluate endpoint flow estimation (LTR) using \textbf{end-point error (EPE)} and \textbf{angular error (AE)}. On DSEC-Flow, we additionally report \textbf{\(N\)-pixel error (NPE)}, i.e., the percentage of valid pixels whose EPE exceeds \(N\) pixels, where \(N\in\{1,2,3\}\), which are also included in the official benchmark metrics.

On MultiFlow, we further evaluate HTR pixel trajectories using \textbf{trajectory end-point error (TEPE)} and \textbf{trajectory angular error (TAE)}, originally proposed by BFlow. TEPE extends EPE by averaging it over \(K\) trajectory timestamps:

\[
\operatorname{TEPE}
=
\frac{1}{K}\sum_{k=1}^{K}
\operatorname{EPE}\!\left(
\mathbf{f}_{\mathrm{pred}}(t_k),
\mathbf{f}_{\mathrm{gt}}(t_k)
\right), \qquad K\geq 2,
\]
and TAE is calculated analogously using AE. Lower values indicate better performance for all metrics.

\input{exps/multiflow}

\subsection{Implementation Details}

\paragraph{MultiFlow}
We train our model for 100 epochs on two NVIDIA V100 GPUs with 32 GB of memory each. We use a batch size of 4 and iteration count $R=5$. We use the AdamW~\cite{loshchilov2017decoupled} optimizer with the OneCycle~\cite{smith2019super} scheduler, setting the maximum learning rate to $1.2\times10^{-4}$, weight decay to $10^{-4}$, and the iterative loss discount factor to $\gamma=0.85$. Gradients are clipped to a maximum norm of $1.0$. The predicted variance for $L_\mathrm{NLL}$ is constrained to $[0,10]$. During training, inputs are randomly cropped to $288\times384$ and augmented with random horizontal and vertical flips. The temporal resolution of input event voxels is set to $K=20$, corresponding to a temporal interval of $25\,\mathrm{ms}$ per voxel.

\begin{figure*}[t]
\centering
  \includegraphics[width=.91\linewidth]{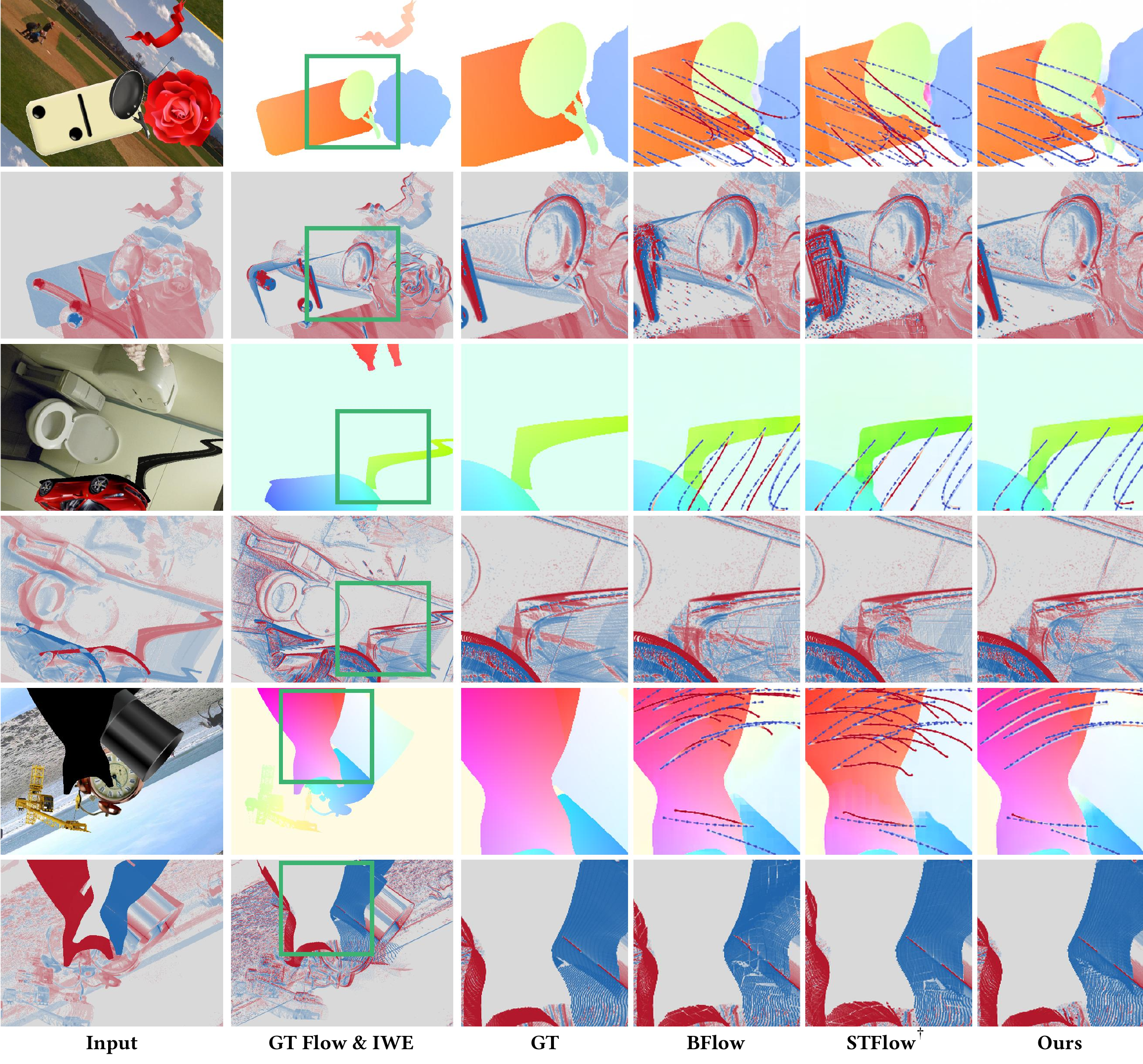}
  \caption{Qualitative results on MultiFlow. For each scene, the first row visualizes the RGB reference image, the ground truth optical flow, and estimated flows from three methods, overlaid with ground truth (blue) and estimated (red) trajectories and zoomed in to a representative region. The second row visualizes the integrated event frame, the ground truth IWE, and IWEs for each method, also zoomed in to the same region.}
  \label{fig:main_comparison}
  \vspace{-3pt}
\end{figure*}

\paragraph{DSEC-Flow}
We make the following changes on the MultiFlow configuration to accommodate for the characteristics of DSEC-Flow. Our models are trained for 100 epochs on a single NVIDIA RTX 5090 GPU with 32 GB of memory. We apply random horizontal and vertical flips without cropping, and use a batch size of 2, while the iterative loss discount factor is set to $\gamma=0.75$. The predicted variance for $L_\mathrm{NLL}$ is constrained to $[0,3]$. We choose $K=8$, which corresponds to a temporal interval of $12.5\,\mathrm{ms}$ per voxel. Finally and most importantly, we set the B\'ezier degree to $L=2$ due to lack of HTR supervision in DSEC-Flow.


\subsection{Experimental Results}

\paragraph{DSEC-Flow} Table~\ref{tab:dsec_results} reports quantitative results on the DSEC-Flow benchmark. As shown, \method achieves comparable performance as previous state-of-art methods under both event only and event-RGB settings. Specifically, under the event-only setting, our method performs slightly better than the newest correlation-free method IDNet~\cite{Wu2024IDNet} and achieve a $8\%$ improvement compared to current best HTR method ResFlow~\cite{zhou2025resflow}. Under the event-RGB setting, our EPE is $7.2\%$ lower than BFlow~\cite{gehrig2024bflow}, thanks to our higher spatiotemporal resolution.
Additionally, our extended models (``+ bidir.'' and ``+ finetuning'') achieve state-of-the-art performance among event-only and event-RGB models, respectively, demonstrating the scalability of our model.


Qualitative results in Fig.~\ref{fig:dsec_compare} further demonstrate how higher feature resolution benefits estimation. By comparing \method against BFlow~\cite{gehrig2024bflow} (event-only checkpoint) and IDNet~\cite{Wu2024IDNet}, operating on 1/8 and 1/4 resolutions respectively, we show that optical flows predicted on lower resolutions tend to be more blurry and ambiguous, sometimes corrupted by patch-like artifacts resulting from the upsampling process. 


\paragraph{MultiFlow} MultiFlow serves as the main benchmark to evaluate HTR optical flow prediction. As shown in Table~\ref{tab:multiflow_results}, our model surpasses other baselines on both trajectory and endpoint metrics under the unimodal setting, beating the best-performing ResFlow~\cite{zhou2025resflow} by $19\%$ in EPE and $20\%$ in TEPE. We observe that most previous methods relying on endpoint correlation can estimate endpoint optical flows reasonably well in the presence of complex and non-linear motions in this challenging dataset, but all LTR models fail significantly on HTR metrics. Notably, IDNet~\cite{Wu2024IDNet} exhibits the worst performance among all baselines, possibly because motion patterns in MultiFlow violates its linear motion assumption.

Under the multimodal setting, \method witnesses a solid performance boost with RGB frames as input. In addition to BFlow and STFlow, we also train STFlow with B\'ezier curve motion representation and B\'ezier-assisted correlation volumes similar to BFlow, which improves the original linear design by a large margin on trajectory metrics, for a more thorough comparison. As reported in Table~\ref{tab:multiflow_results}, \method outperforms both versions of STFlow and BFlow in trajectory metrics, and it achieves similar endpoint accuracy to the B\'ezier-enhanced STFlow model. While the original STFlow ranks first in endpoint prediction, its trajectory estimation is way worse than other models.
Finally, we train the RGB-only WAFT model to probe the influence of event information on scene dynamics modeling. We find that WAFT achieves competitive endpoint metrics without any motion cue between two frames, but it is still outperformed by our model, revealing that event streams carry critical information that helps to capture object motion more accurately. 

Figure~\ref{fig:main_comparison} visually compares \method against BFlow and STFlow\textsuperscript{\textdagger} from three aspects: optical flow, pixel trajectory, and image of warped events (IWE). Specifically, IWEs are obtained by warping events to the reference timestamp $\tau_0$ using predicted (or ground truth) continuous optical flows and integrating them together, which serves as an indicator of the alignment between flow maps and event streams. In all three sampled instances, our method achieves the best trajectory alignment.

\input{exps/ablation_traj}

\input{exps/ablation_ST}

\subsection{Ablation Study}
We conduct ablation experiments for warping policies and trajectory loss on MultiFlow following the event-only settings and report our results in Table~\ref{tab:ablation}. Additionally, we also investigate how model performance changes with spatial and temporal resolutions, which validates our claim that \method gains advantage by scaling up to higher spatiotemporal resolutions.

\paragraph{Trajectory-based warping.}

We compare three different warping policies: degree-10 B\'ezier curve warping, linear warping (realized by setting B\'ezier degree to 1), and no warping. We observe that linear warping policy performs even worse than the no warping policy on both trajectory and endpoint metrics, while both of them fall behind the B\'ezier policy, demonstrating the necessity of representing pixel trajectory and conducting feature warping at a sufficiently high degree of freedom to model highly nonlinear motion.

\begin{figure*}[ht]
    \centering
    \includegraphics[width=0.99\linewidth]{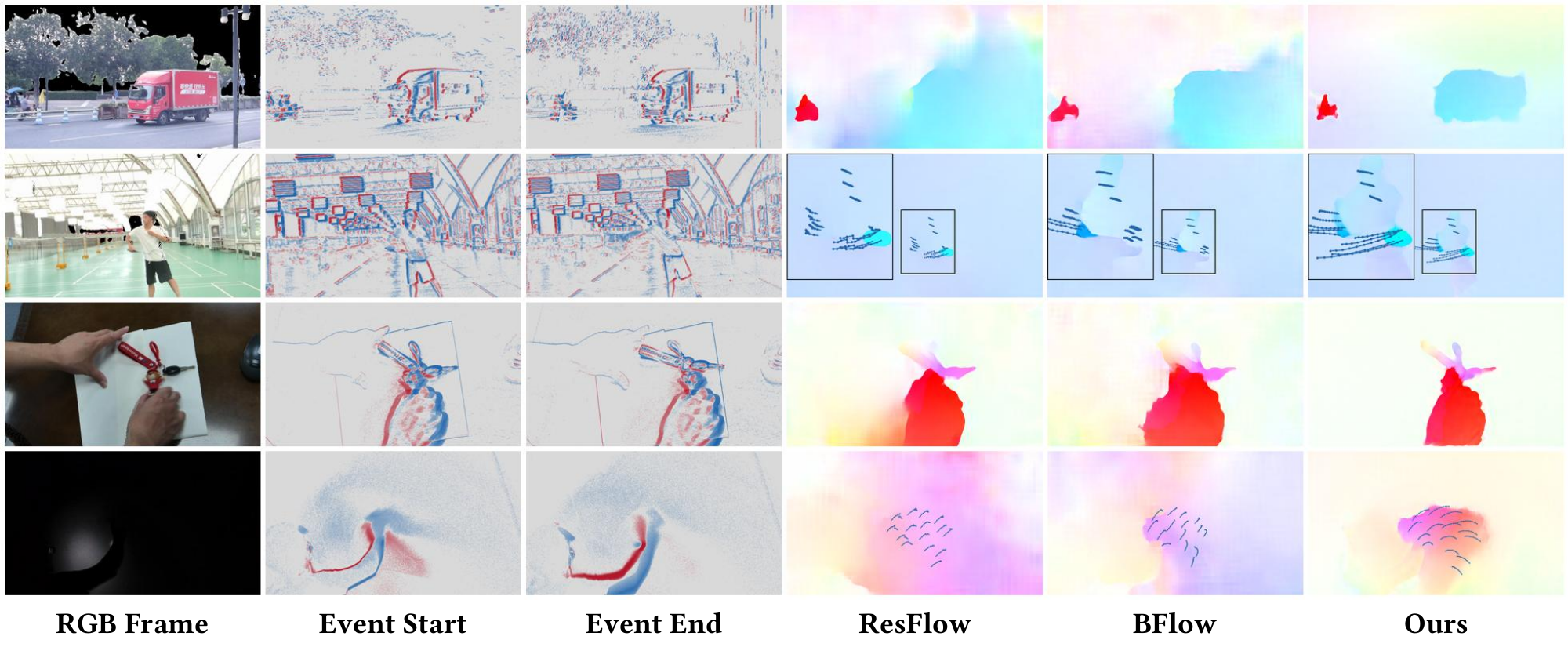}
    \caption{Results on real-world acquisitions. From left to right: reference RGB frames (not used for inference), event voxels at the beginning and end of inference window, and estimated results by three methods. For scenarios 2\&4, we additionally visualize pixel trajectories (blue) to intuitively compare motion estimation accuracy across models.}
    \label{fig:real_world_performance}
    
\end{figure*}

\paragraph{Trajectory loss.}

We further ablate with the trajectory loss $L_{\text{traj}}$ by setting  $\lambda_{\text{traj}}$ to $0$. Results in Table~\ref{tab:ablation} show that degree-10 B\'ezier curves are highly underdetermined without dense trajectory supervision, which results in a severe degradation of prediction accuracy, especially in TEPE and TAE. However, when the linear warping policy is in effect, removing the trajectory loss leads to better endpoint prediction since the extra supervision between endpoints serves only as a noise factor in this case.

\paragraph{Spatial Resolution}

We adjust the patch size of the DPT head to realize spatial resolution tuning without drastically changing the number of parameters. For example, spatial resolution is set to $1/4$ and $1/8$ by choosing patch sizes $4\times4$ and $2\times2$, respectively. As shown in Table~\ref{tab:ablation_st}, performance consistently degrades as spatial resolution decreases, demonstrating the importance of maintaining a high spatial resolution to preserve fine-grained spatial information.

\paragraph{Temporal Resolution}

Temporal resolution is characterized by the number of event voxels used for prediction, which reflects how densely the event stream is sampled over time. For all experiments, we keep the temporal duration of each voxel fixed at $25$\,ms. In Table~\ref{tab:ablation_st}, we observe that increasing the voxel count from 5 to 20 consistently improves all metrics. This trend shows that a denser temporal representation provides a more comprehensive description of intermediate motion between RGB timestamps.

\subsection{Real-world Assessment}

To assess real-world applicability of different models, we conduct zero-shot qualitative evaluations on self-captured data from diverse indoor and outdoor scenes, where ground truth motion is unavailable. All models are event-only and trained on the MultiFlow dataset. This Sim2Real setting better aligns with actual use cases where fine-tuning on large-scale labeled data is often infeasible.

We construct a rigid head-mounted prototype comprising a Prophesee EVK4 event camera and an Orbbec Femto Bolt RGB-D camera. EVK4 records events at $1,280\times720$ resolution with $\mu$s timestamps, while our Femto Bolt capture pipeline provides viewpoint-aligned $1,920\times1,080$ RGB-D frames at 30 FPS. We use a hardware trigger to synchronize each RGB-D frame with the corresponding event stream segment. For visualization, we refine depth maps with the pretrained LingBot-Depth v0.5 model~\cite{lingbot-depth2026} and project RGB images onto the event camera viewpoint using depth maps and calibrated RGB-to-event geometry.

Figure~\ref{fig:real_world_performance} visualizes our results. \textbf{Rows 1--2} are captured in bright outdoor scenarios with both camera egomotion and object motion. 
In the ``streetview'' scenario, \method produces clearer and more stable motion boundaries for the moving truck and pedestrians. In the ``badminton'' scenario, where fast-moving objects (right arm and racket) can lead to motion blur for the frame-based RGB camera, our method still performs stably and estimates trajectories that are more realistic and coherent than ResFlow and BFlow, both state-of-the-art HTR models.

\textbf{Rows 3} feature indoor captures and challenge each method's ability to track and recognize hand and small objects (\textit{e.g.,} key chain), which move along highly nonlinear patterns. Thanks to the high spatial resolution of \method, our model is able to make fine-grained predictions such as distinct boundaries for fingers, while correlation volume-based methods are restricted by their designs and can only give fuzzy estimates.

\textbf{Row 4} is captured under extreme low-light condition, where illumination is restricted solely to a smartphone flashlight. As shown in columns 1--3, the RGB camera fails to capture usable scene detail, whereas the event camera still records valuable visual information. Nevertheless, low-light event streams are sparser and more noisy, making optical flow estimation even more challenging. Columns 4--6 show that \method not only recovers object boundaries (an arm wiping the blackboard), but also infer plausible trajectories for pixels invisible to RGB cameras.  


%% file: exps/dsec.tex
\begin{table}
\centering
\footnotesize
\caption{Quantitative results on DSEC-Flow. Methods up to ``Ours'' are event-only, while STFlow, BFlow and ``Ours w/ img'' works on event-RGB inputs. ``+ bidir.'' and ``+ finetuning'' correspond to two extensions in Sec.~\ref{sec:extension}, respectively.}
\begin{tabular}{l|llllll} 
\hline\hline
Method  & 1PE & 2PE & 3PE & EPE & AE & HTR  \\ 
\hline
E-RAFT   & 12.74    &   4.74  &  2.68   &  0.79   &   2.85 &      \\
TMA     &  10.86   &   3.97  &  2.30   &  0.74   &  2.68  &      \\
IDNet   & 10.07    &  3.50   &  2.04   &  0.72   &  2.72  &      \\
ECDDP   &  8.89 & 3.20	&  1.96	 &   0.70 &  2.58 &      \\
BAT     &  7.54 & 2.84  & 1.74 &  0.65 &  2.43 &      \\
ResFlow &  11.22   &  4.24   &  2.50   & 0.75   &  2.73  &   \checkmark    \\
Ours    &  8.82   & 3.33 & 2.07 & 0.69    &  2.52  &   \checkmark    \\
\hspace{4pt}  + bidir. & 7.70 & 2.88 & 1.75 & 0.63 & 2.37  &\checkmark \\
\hline
STFlow  & 7.93  &  	2.61  &  	1.45  &  0.63&   2.29     &      \\
BFlow   & 9.70  &  3.42   & 1.88    & 0.69    &  2.42  & \checkmark     \\
Ours w/ img   & 8.15    &   2.83  &   1.73  &   0.64  &  2.40  &   \checkmark    \\
\hspace{4pt} + finetuning &  7.26&  2.54&  1.56& 0.62 & 2.24  &\checkmark  \\
\hline\hline
\end{tabular}
\label{tab:dsec_results}
\end{table}

%% file: exps/multiflow.tex
\begin{table}
\centering
\footnotesize
\caption{Quantitative results on MultiFlow. STFlow\textsuperscript{\textdagger} denotes STFlow model trained with Bezier curve motion representation.}
\arrayrulecolor{black}
\begin{tabular}{ll|lllll} 
\hline\hline
Methods & Input & TEPE & TAE   & EPE  & AE    & HTR  \\ 
\hline
WAFT          & i        & 6.20 & 17.85 & 3.85 & 3.36  &      \\
IDNet         & e        & 7.08 & 20.74 & 7.60 & 10.85 &      \\
E-RAFT        & e        & 6.21 & 18.74 & 4.35 & 5.7   &      \\
TMA           & e        & 6.02 & 18.24 & 3.91 & 4.84  &      \\
BFlow         & e        & 2.74 & 7.00  & 5.00 & 7.35  & \checkmark    \\
ResFlow       & e        & 2.28 & 5.32  & 3.74 & 4.70  & \checkmark    \\
Ours          & e        & 1.81 & 3.89   &  3.03 &3.51 & \checkmark    \\ 
\hline
STFlow        & e+i      & 5.16 & 16.67 & 1.65 & 1.99  &      \\
STFlow\textsuperscript{\textdagger} & e+i & 1.65 & 4.86  & 1.98 & 2.56  & \checkmark    \\
BFlow        & e+i      & 2.05 & 5.33 & 3.55 & 5.00 & \checkmark    \\
Ours         & e+i      & 1.25  & 2.95 & 2.04 & 2.45 & \checkmark    \\
\hline\hline
\end{tabular}
\label{tab:multiflow_results}
\end{table}

%% file: exps/ablation_traj.tex
\begin{table}
\centering
\footnotesize
\caption{Ablation study results. ``B" denotes warping with degree-10 B\'ezier curves, ``L" denotes warping with linear trajectories, ``no" denotes prediction without warping process}
\label{tab:ablation}
\begin{tabular}{cc|cccc} 
\hline\hline
\multicolumn{2}{c|}{Methods} & \multicolumn{2}{c|}{Trajectory} & \multicolumn{2}{c}{Endpoint}  \\
$L_{\text{traj}}$ & Warp                    & TEPE & \multicolumn{1}{c|}{TAE}    & EPE & AE                                      \\ 
\hline
 \checkmark  & B  &         1.81    &   3.89   &  3.03 &  3.51                                            \\
 $ \times $    &   B &       7.62   &  24.60   &   4.87 &5.34                                      \\

  \checkmark  &     L &   4.64   &     14.64  & 6.84    &   7.07                                       \\
 $ \times $    &     L  &    6.27   &    18.32 & 4.70  & 5.14                                        \\
 \checkmark & no & 3.21 &6.96 &  5.58 & 6.51 \\
\hline\hline
\end{tabular}

\end{table}

%% file: exps/ablation_ST.tex
\begin{table}
\centering
\footnotesize
\caption{Ablation study results on spatial and temporal resolutions. Spatial resolutions are written with respect to the original data resolution, and temporal resolution is denoted by the number of event voxels used for prediction.}
\label{tab:ablation_st}
\begin{tabular}{cc|cccc} 
\hline\hline
\multicolumn{2}{c|}{Resolution} & \multicolumn{2}{c|}{Trajectory} & \multicolumn{2}{c}{Endpoint}  \\
Spatial & Temporal                    & TEPE & \multicolumn{1}{c|}{TAE}    & EPE & AE                                      \\ 
\hline
 1/2  & 20  &  1.81    &   3.89   &  3.03 &  3.51                                \\
 1/4    &   20 & 2.02 & 4.73  &3.39  & 4.28                                \\

 1/8  &     20 & 2.46 & 6.09  & 3.91  & 4.79                                    \\
\hline
 1/2  & 20  &  1.81    &   3.89   &  3.03 &  3.51      \\
 1/2     &    10  &  2.05    & 4.46    & 3.39  & 3.98                                       \\
  1/2   &    5 & 2.22  & 4.65&3.71  &4.18  \\
\hline\hline
\end{tabular}
\end{table}

%% file: sections/conclusion.tex
\section{Discussions and Conclusion}
\label{sec:discussion_and_conclusion}

\paragraph{Limitations and Future Work.}

Our work has several limitations. Firstly, it relies on dense event voxel encoding, whose computational cost scales with the input event frame rate. Future research could incorporate more light-weighted encoders~\cite{dalgaty2023hugnet} or sliding-window approaches~\cite{Gehrig24nature}.
Secondly, our model trains on ground truth annotations, but it is potentially generalizable to a hybrid-supervision framework, following STSC-Flow~\cite{hu2026stsc} and ResFlow~\cite{zhou2025resflow}. 
Finally, real-world event acquisition suffers from various noise factors such as flickering~\cite{han2025edef}, reflection~\cite{wang2026evreflection}, and shadow. 
These artifacts can degrade flow accuracy in unconstrained environments, indicating that targeted noise-robustness mechanisms are necessary prior to deployment in VR/AR devices.

\paragraph{Conclusion.}
We present \method, a correlation-free framework for high temporal resolution (HTR) optical flow estimation from event streams. By replacing all-pairs correlation with the attention mechanism with warping-aligned features, \method effectively exploits continuous temporal information inherent in event data while maintaining its scalability to higher spatiotemporal resolutions. 
Extensive evaluations on DSEC-Flow and MultiFlow show that \method achieves state-of-the-art performance in HTR motion estimation, as well as LTR estimation with minor modifications. 
Real-world assessments using a head-mounted prototype further confirm its robustness under severe underexposure as well as rapid, non-linear motion dynamics.
These results present \method as a promising candidate for facilitating temporally dense, dynamic perception in immersive VR/AR systems.